\documentclass{article}

\usepackage{PRIMEarxiv}

\usepackage[utf8]{inputenc} 
\usepackage[T1]{fontenc}    
\usepackage{hyperref}       
\usepackage{url}            
\usepackage{booktabs}       
\usepackage{amsfonts}       
\usepackage{nicefrac}       
\usepackage{microtype}      
\usepackage{lipsum}
\usepackage{fancyhdr}       
\usepackage{graphicx}       
\usepackage{booktabs}
\usepackage{url}
\usepackage[table,xcdraw]{xcolor}
\graphicspath{{media/}}     

\title{Creating a Safety Assurance Case for a Machine Learned Satellite-Based Wildfire Detection and Alert System
}

\author{
  \href{https://orcid.org/0000-0001-7347-3413}{\includegraphics[scale=0.06]{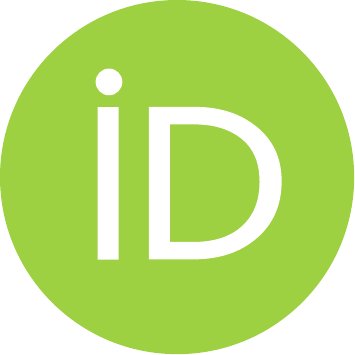}\hspace{0.5mm}Richard Hawkins}\thanks{Corresponding author: Dr Richard Hawkins, Department of Computer Science,
        University of York, York, UK. \textit{email: richard.hawkins@york.ac.uk}}, Chiara Picardi \\
    Department of Computer Science \\
  University of York \\
  York, UK\\
  \texttt{\{richard.hawkins|chiara.picardi\}@york.ac.uk} \\
\AND
Lucy Donnell, Murray Ireland\\
Craft Prospect Ltd\\
\texttt{\{lucy|murray\}@craftprospect.com} \\
}

\begin{document}
\maketitle

\begin{abstract}

Wildfires are a common problem in many areas of the world with often catastrophic consequences. A number of systems have been created to provide early warnings of wildfires, including those that use satellite data to detect fires. The increased availability of small satellites, such as CubeSats, allows the wildfire detection response time to be reduced by deploying constellations of multiple satellites over regions of interest. By using machine learned components on-board the satellites, constraints which limit the amount of data that can be processed and sent back to ground stations can be overcome. There are hazards associated with wildfire alert systems, such as failing to detect the presence of a wildfire, or detecting a wildfire in the incorrect location. It is therefore necessary to be able to create a safety assurance case for the wildfire alert ML component that demonstrates it is sufficiently safe for use. This paper describes in detail how a safety assurance case for an ML wildfire alert system is created. This represents the first fully developed safety case for an ML component containing explicit argument and evidence as to the safety of the machine learning.

\end{abstract}

\keywords{Machine learning \and Safety case \and Safety assurance \and Wildfire \and Satellite}

\section{Introduction}

Wildfires are a common and often catastrophic occurrence in many parts of the world. In the 2019 and 2020 Australian bushfire season, over 18 million hectares of forest and 10,000 buildings were destroyed and six people killed \cite{morton2019}. The Gangwon wildfire in South Korea in 2019 burnt 500 hectares of land and destroyed several hundred buildings \cite{bbc2022}. 
2020 and 2021 were the worst years for wildfires in the USA in at least 10 years with an average of 69,000 wildfires burning over 6 million acres each year, based on data compiled by the National Interagency Fire Center (NIFC).  As well as loss to life, the financial cost of these wildfires is immense, so effectively managing the prevention and response to wildfires is crucial. Early detection of emerging wildfires enables them to be suppressed and managed, reducing the requirement for costly and dangerous firefighting.

There are three types of system used for wildfire detection: terrestrial, airborne, and spaceborne systems \cite{barmpoutis2020review}. In this paper we focus on spaceborne wildfire detection. Services such as the Fire Information for Resource Management System (FIRMS) \cite{firms20}, the Global Wildfire Information System (GWIS) \cite{gwis} and the Copernicus Emergency Management System (EMS) \cite{copern} have been created to provide early warnings, statistical data and coverage maps for wildfires. Such services rely heavily on satellite data to provide the perspective, spectral content and temporal frequency needed for regular and accurate detection and reporting of wildfires. As these services rely on existing satellite missions, however, they are subject to the limitations of these missions in terms of visit frequency, information latency and quality of data. For example, FIRMS reports a lead time of 3 hours from observation (not the fire actually starting or being observable) to reporting on the ground \cite{firms20}, a geolocation precision of 375 m \cite{viirs} or 1 km \cite{mcd14} and a false positive error of 1.2\% \cite{billing83}. The source satellites used for FIRMS (Terra, Aqua, Suomi NPP and NOAA-20) have a revisit time of between 14 hours and 2 days. This makes the worst-case scenario for a detection response time around 51 hours, assuming a fire becomes observable immediately following a satellite pass. While emergency services do not rely exclusively on platforms such as FIRMS, the ability to provide warnings even a few hours earlier could make a huge difference to the preservation of human, animal and plant life and infrastructure.

 The detection response time on fire alerts can be reduced by increasing the revisit frequency of the satellites or deploying a constellation that is intentionally sized and designed to meet specific revisit and latency requirements. This has become possible with the increased availability of space assets, such as CubeSats (the most popular form factor for small satellites). There are, however, constraints on resources such as power and bandwidth when using CubeSats which limit the amount of data that can be processed and sent back to ground stations. In the case study we consider in this paper, such bottlenecks are overcome by using machine learned (ML) components on-board the satellites to detect wildfires and generate timely and data-efficient alerts which are transmitted to a ground station. The response authority (such as the fire service) considers the alerts and determines an appropriate response based on a number of factors such as the number of fires detected in a specific catchment area, the distribution of the fires and distance from both each other and the response team’s base.

There are potential hazards associated with a wildfire alert system such as this. Failure to detect the presence of a wildfire or detecting a wildfire in the incorrect location could lead to a delay in the response to the fire, a larger and less controlled fire, and thus potentially increasing the risk of harm to people and property or putting firefighting teams in danger. Conversely, raising an alert for a wildfire that doesn't actually exist could result in fire response resource being mis-assigned and thus unavailable to respond to real wildfires in a timely manner. It is necessary therefore to be able to provide confidence in the alerts generated by the satellite-based fire detection system such that they can be trusted.  To do this, for the ML component that is used for wildfire detection and alerting, we need to create a safety assurance case that presents a compelling argument that the component is sufficiently safe, supported by rigorous evidence. 

In this paper we describe in detail how a safety assurance case for an ML wildfire alert system is created. The paper is structured as follows. Section 2 discusses safety cases for ML software and how they can be created. Section 3 provides a description of the wildfire alert system. The safety case is presented in section 4. Section 5 provides conclusions and discusses future work directions.

\section{Safety Assurance Cases for Machine Learning}\label{sec:safetyCase}

\begin{figure}[h]
    \centering
    \includegraphics[width=1\linewidth]{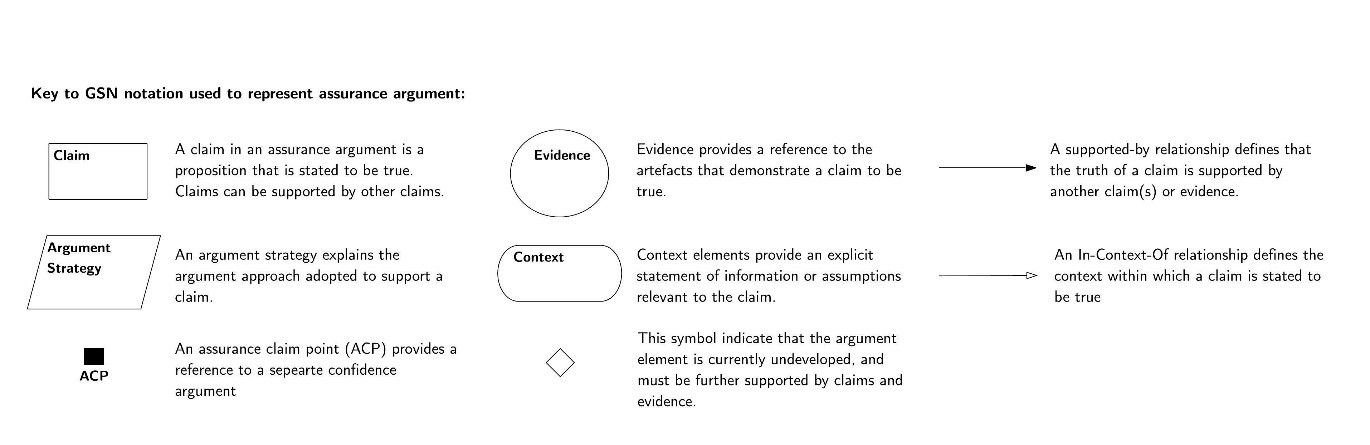}
    \caption{Key to GSN notation}
    \label{fig:gsnKey}
\end{figure}

In order to demonstrate that a system is acceptably safe to operate, it is common to provide a safety case for that system.
A safety case comprises “a structured argument, supported by a body of evidence, that provides a compelling, comprehensible and valid case that a system is safe for a given application in a given environment” \cite{m2017a}. For systems that contain software, the safety case must consider the contribution of the software to the safety of the overall system. Creating a an explicit safety case containing a structured argument and evidence helps to provide explicit safety justification, making
it easier to understand, review and criticise the reasoning and evidence presented. One approach that is commonly used to present the safety arguments for a safety case is the Goal Structuring Notation (GSN) \cite{group2018a}. The basic elements of GSN are shown in Figure \ref{fig:gsnKey}. 

These GSN elements can be used to construct a safety argument by showing how safety claims are broken down into sub-claims, until eventually they can be supported by evidence. The strategies adopted, and the rationale (assumptions and justifications) can be captured, along with the context in which the goals are stated. Confidence arguments relating to various aspects of the safety case can be provided. Assurance claim points (ACPs) can be used to indicate where such arguments are provided. In this paper GSN is used to present the safety arguments.

Previous work has been undertaken looking at how to develop safety cases for safety-related software systems, such as \cite{hawkins2013assurance} and in a number of domains, standards require the production of a safety case for software elements of a system \cite{m2017a}, \cite{iso2018}. However this previous work has focused on traditional software and not considered machine learning. These existing software safety assurance approaches do not apply well to ML software for a number of reasons including:

\begin{enumerate}
    \item They assume a development process based around the decomposition of requirements down to the level of implementation.
    \item They assume the software generated can be understood and analysed by humans.
    \item They assume that defined test coverage metrics can be used to judge the sufficiency of the testing undertaken.
\end{enumerate}

\begin{figure}[h]
    \centering
    \includegraphics[width=1\linewidth]{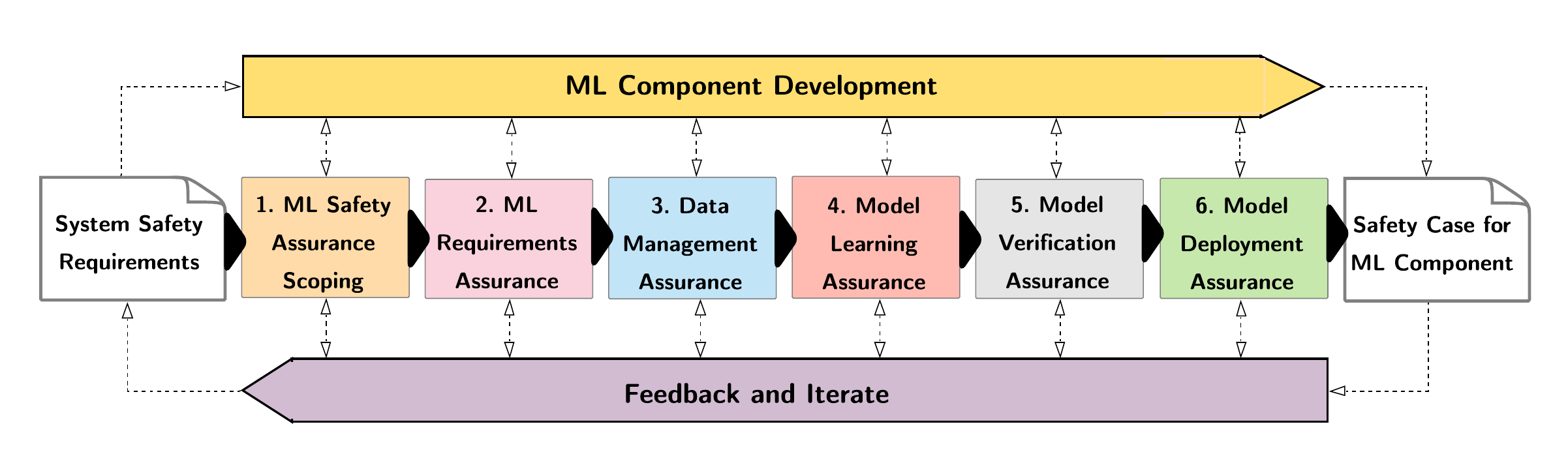}
    \caption{Overview of the AMLAS Process}
    \label{fig:amlas}
\end{figure}

None of these assumptions hold for ML software, where a completely different development approach is adopted, the resulting software algorithm is opaque to human interpretabilty, and traditional coverage metrics are meaningless. In response to this the authors, in previous work, have developed an approach for safety assurance of machine learning (AMLAS) \cite{hawkins2021guidance}. AMLAS is a process that consists of 6 stages, as shown in figure \ref{fig:amlas}. For each stage the AMLAS process describes a set of activities that can be followed, and the artefacts that are generated. It then details how these artefacts may be used to create a safety case for the ML component. In section \ref{sec:system} we apply each stage of the AMLAS process to a satellite-based wildfire alert ML component to create a compelling safety case.

\section{Wild Fire Alert System Description}\label{sec:system}

\begin{figure}[h]
    \centering
    \includegraphics[width=0.8\linewidth]{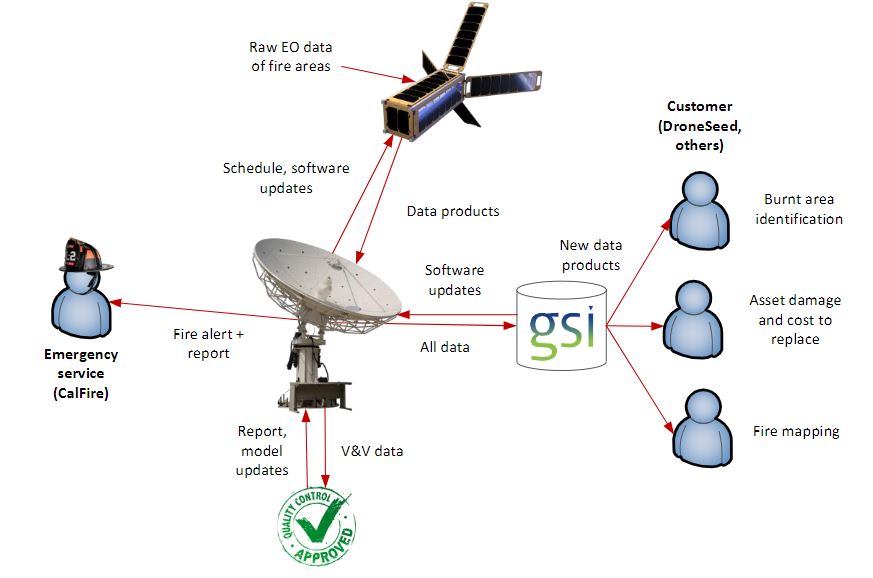}
    \caption{Concept of Operations for Wildfire Alert System}
    \label{fig:dataFlow}
\end{figure}

The concept of operations for the detection system is shown in figure \ref{fig:dataFlow}.  A satellite with a multi-spectral imager passes over a region of interest that may contain wildfires. The imager operates on a set frequency, capturing images of the sub-satellite environment and classifying them using a neural network trained on satellite images of fires in the spectrum of the imager. The neural network detects the presence of any fires in the image, and transmits a lightweight  text  alert  containing the location of the fire and time of detection to the groundstation. The fire alerts are prioritised and downlinked  to  the  ground  ahead  of  all  other  data. This alert is then passed to the response authority.

In order to maximise the time during which a satellite is available to obtain images of a particular area of interest, multiple CubeSats are used for this application. 8 standard  6U  platforms are employed,  each hosting  identical instrument payloads and subsystems. The orbit of the satellites and their instruments will reflect those of Sentinel-2 and Landsat 8, which are the sources of the training data for the ML component. The  satellites are in  a  sun-synchronous  low  Earth  orbit (LEO) at 450km  altitude  and  97.2° inclination. They orbit the Earth approximately every 94 minutes and are evenly distributed around the ascending node, such that revisit times between satellites for a given location are constant. The satellites use a generic 6U bus with standard ADCS components including a 9 DOF IMU, coarse and fine sun sensors, reaction wheels and magnetorquers. They are capable of fine pointing at specific LLA targets or the satellite nadir and velocity vectors.

\begin{figure}[h]
    \centering
    \includegraphics[width=1\linewidth]{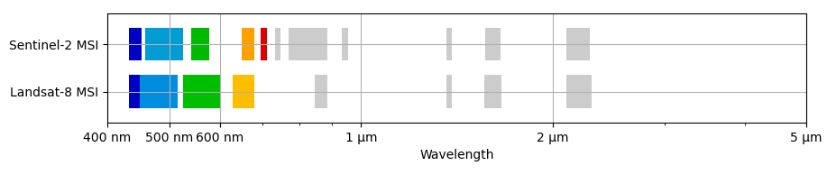}
    \caption{Spectral bands for Sentinel-2 and Landsat-8 MSIs}
    \label{fig:payload}
\end{figure}

The satellite payload comprises a generic multispectral instrument (MSI) which is similar to the MSIs used  on  Sentinel-2 and  Landsat-8. The  bands  of  the  instrument are also common to both Sentinel-2 and Landsat-8, shown in Figure \ref{fig:payload}. 

The MSI has the following properties:
\begin{itemize}
    \item Ground footprint: 32.5 x 19.6 km
    \item Max ground resolution: 10 m/px
\end{itemize}

\begin{figure}[h]
    \centering
    \includegraphics[width=0.3\linewidth]{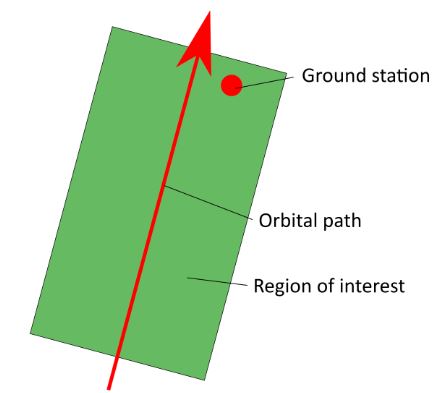}
    \caption{Ground station location in region of interest}
    \label{fig:roi}
\end{figure}

A single ground station is used, which will be located at the far end of the region of interest (RoI) with respect to the direction of travel of the satellite as shown in figure \ref{fig:roi}. This ensures that fire alerts in the RoI are downlinked as soon as possible after identification.

Figure \ref{fig:dataFlow} also indicates how the satellite can be used for other commercial applications by providing larger data products to commercial customers such as burnt area identification and asset damage information as well as more detailed fire mapping. This may include sending full images to the groundstation. These applications require more data processing and transmission and therefore take longer than the prioritised fire alerts, however since these are commercial use cases that have no direct safety impact they are not time-critical in the same way as the fire alerts. These commercial applications are not considered in this paper.

Figure \ref{fig:dataFlow} also indicates that verification of the on-board fire detection can be performed on the ground during operation through (non-real-time) verification against groundtruth data from other fire detection sources. Where necessary this verification could lead to software updates to improve the operational performance of the ML component.

\section{The Safety Case for the Wild Fire Alert ML Component}\label{sec:SC}

\subsection{ML Assurance Scoping} \label{sec:scope}

The objectives at this first stage are to define the scope of the safety case and of the safety assurance process for the wildfire alert ML component. This stage establishes the top‐level safety assurance claim of the safety case and specifies the relevant contextual information for the ML safety argument. Since the safety of the ML component cannot be assured in isolation from the broader wildfire alert system, this stage ensures the assurance of the ML component takes account of the overall system and the system-level safety process.

There are a number of key artefacts that are required for this stage of the safety case. This includes the documented descriptions of the system and the operating environment as summarised above. In addition, the system safety requirements for the wildfire alert system must be specified. These safety requirements were generated from following a system safety assessment process, the details of which are outside of the scope of this paper. The system safety assessment process identified 2 hazards for the wildfire alert system as shown below. Against each hazard a number of safety requirements were defined in order to manage those hazards as detailed in Table \ref{tab:SRs}.







\begin{table}[]
\caption{\label{tab:SRs}System safety requirements for wildfire alert system}
\begin{tabular}{@{}ll@{}}
\toprule
\multicolumn{2}{l}{\textbf{Hazard 1 - Services Miss an Emergency}}                                                                                                                                                                               \\ \midrule
\multicolumn{1}{l|}{\textbf{REQ-SAFE-ER-1}} & \begin{tabular}[c]{@{}l@{}}The Emergency Response Service shall determine the location\\  of an active wildfire within 200 m of its true location.\end{tabular}                                    \\ \midrule
\multicolumn{1}{l|}{\textbf{REQ-SAFE-ER-2}} & \begin{tabular}[c]{@{}l@{}}The Emergency Response Service shall inform emergency \\ services of an active wildfire with 3 hours of it starting.\end{tabular}                                       \\ \midrule
\multicolumn{1}{l|}{\textbf{REQ-SAFE-ER-3}} & \begin{tabular}[c]{@{}l@{}}The Emergency Response Service shall positively identify 95\% of \\ all active wildfires acquired by the satellite instrument within the area of interest.\end{tabular} \\ \midrule
\multicolumn{2}{l}{\textbf{Hazard 2 - Services are Directed to a False Emergency}}                                                                                                                                                               \\ \midrule
\multicolumn{1}{l|}{\textbf{REQ-SAFE-ER-4}} & \begin{tabular}[c]{@{}l@{}}The Emergency Response Service shall falsely indicate active \\ wildfires in the area of interest at a rate not exceeding current fire alert service.\end{tabular}      \\ \bottomrule
\end{tabular}

\end{table}


The responsibility for satisfying each of these system level safety requirements lies with multiple elements of the overall system such as the satellite itself and its sensing and hardware components, the ground station and its components, the communication links between the elements, and so on. The safety case for the overall system considers the assurance of all of these elements, including their integration and interaction. This overall system safety case for the wildfire alert system is outside of the scope of this paper.

Some of the responsibility for assuring that the system safety requirements are met can also however be seen to lie with the ML component onboard the satellite. Specifically, requirements 1, 3 and 4 above can be partly allocated to the ML component\footnote{Satisfaction of safety requirement 2 is dominated by factors such as the re-visit time of the satellites and communication times. As such it does not need to be allocated to the ML component}. It is important to note here however that at this stage there is nothing in these safety requirements that relates in particular to ML. These safety requirements represent what the component is required to do in order to be safe, and the requirements could equally apply to a traditional (non-ML) component if that was being used instead. These system safety requirements were turned into specific ML requirements later in the AMLAS process.

\begin{figure}[h]
    \centering
    \includegraphics[width=0.8\linewidth]{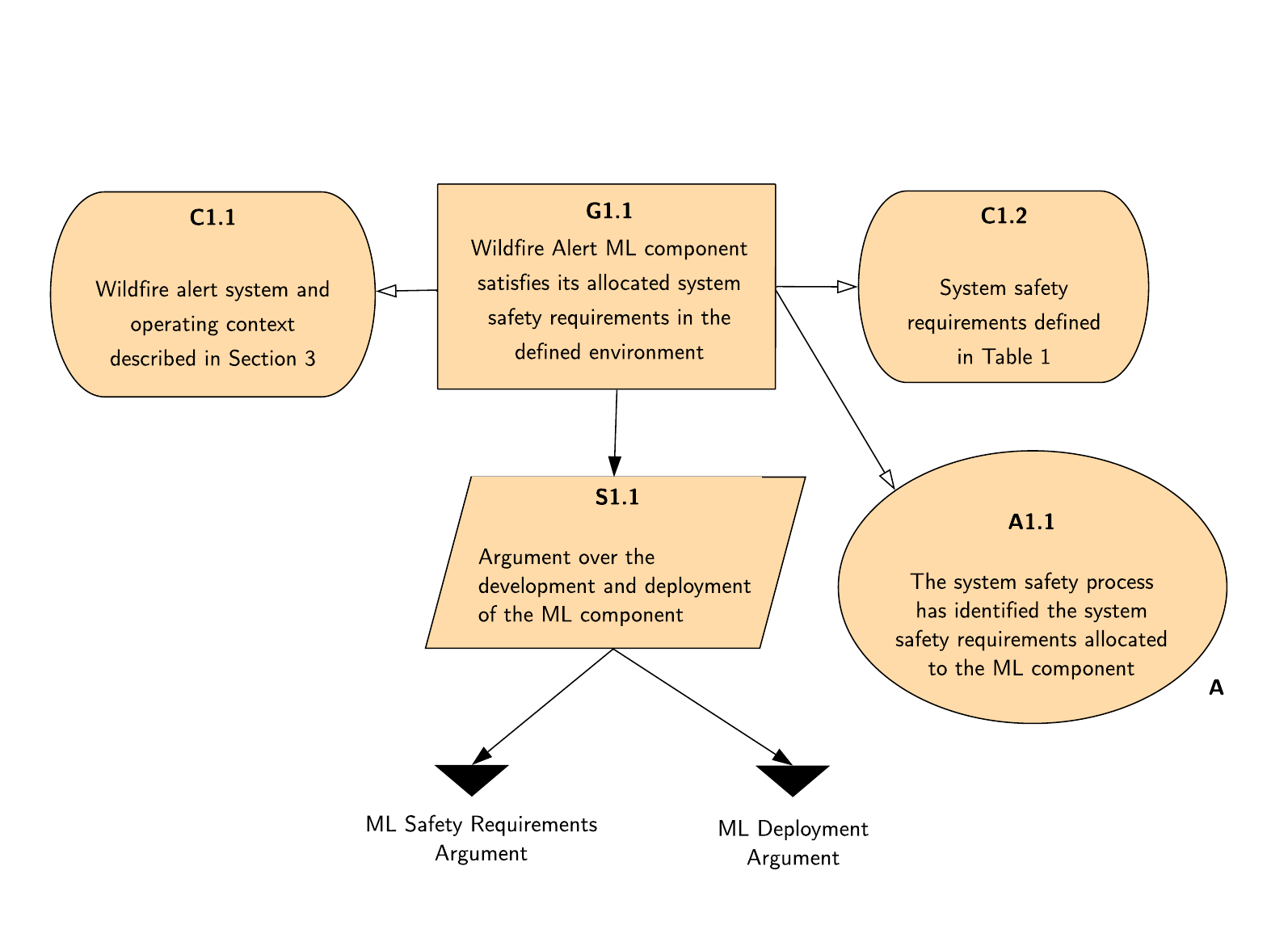}
    \caption{ML Assurance Scoping Argument for wildfire alert component}
    \label{fig:scopeArg}
\end{figure}

As for all stages of the AMLAS methodology, the artefacts discussed above were then used to create the relevant part of the safety argument for the wildfire alert ML component as shown in figure \ref{fig:scopeArg}. The argument explicitly lays out the system safety requirements that the ML component must satisfy (C1.2), as well as clearly scoping both the system and operating context for which the safety case is valid (C1.1). The safety argument also explicitly states the assumption upon which the safety case for the ML component relies (A1.1), which is that the system safety process has correctly identified the system safety requirements. The validity of this assumption is demonstrated as part of the overall system safety case (not shown here).

It can be seen in figure \ref{fig:scopeArg} that this top-level safety claim for the ML component is supported by further argument and evidence from the other stages of the AMLAS process (the ML safety requirements argument and the ML deployment argument) discussed in the sections \ref{sec:MLSR} and \ref{sec:deploy}.
\subsection{ML requirements Assurance}\label{sec:MLSR}

The next stage of the process takes the system safety requirements that relate to the ML fire alert component that were defined at the previous stage and from those, derives a set of specific \emph{ML} safety requirements. This requires that the informal, technology-agnostic safety requirements that have already been identified are translated into a format, and a level of detail that is amenable to ML implementation and verification. The definition of the ML safety requirements must take account of the concept of operations of the wildfire alert system and the overall system and operating context described at the previous stage. 

The ML safety requirements include requirements for performance and robustness of the ML model. We present in Table \ref{tab:MLSRs} each of the ML safety requirements that was specified for the wildfire alert ML component. In this case the robustness requirement is defined with respect to a set of classes. Table \ref{tab:Rob} provides each of these classes. Any values for each class that were determined not to be in scope for the ML component in this particular application are indicated in the table with an `x' in the final column.

\begin{table}[]
\caption{\label{tab:MLSRs}ML safety requirements for wildfire alert system ML component}
\begin{tabular}{@{}ll@{}}
\toprule
\multicolumn{2}{l}{\textbf{Performance}}                                                                                                                                                                            \\ \midrule
\multicolumn{1}{l|}{\textbf{MLSR1}} & \begin{tabular}[c]{@{}l@{}}All points of the mask generated by the ML component shall be less than 6 pixels outside the \\ boundary of the area of the real fire. \end{tabular}                                    \\ \midrule
\multicolumn{1}{l|}{\textbf{MLSR2}} & \begin{tabular}[c]{@{}l@{}}The ML component shall correctly identify the presence of a fire that satisfies the Schroeder \cite{schroeder2016active}\\ conditions  in a frame for 95\% of real fires. \end{tabular}                                       \\ \midrule
\multicolumn{1}{l|}{\textbf{MLSR3}} & \begin{tabular}[c]{@{}l@{}}The ML component shall not identify the presence of a fire in a frame where there is not a real\\ active fire more than 52 times per month.\end{tabular} \\ \midrule
\multicolumn{2}{l}{\textbf{Robustness}}                                                                                                                                                               \\ \midrule
\multicolumn{1}{l|}{\textbf{MLSR4}} & \begin{tabular}[c]{@{}l@{}}ML performance requirements shall be satisfied for all data across the range of classes identified in\\ Table \ref{tab:Rob}.\end{tabular}      \\ \bottomrule
\end{tabular}

\end{table}

\begin{table}[]
\caption{\label{tab:Rob}Relevant Robustness Classes for Wildfire Alert Model}
\begin{center}
\begin{tabular}{|l|l|l|}
\hline
\rowcolor[HTML]{C0C0C0} 
\textbf{Element} & \textbf{Value}                                                        & \textbf{In-context?} \\ \hline
Land   type      & Temperate rainforest                                                  &                      \\ \cline{2-3} 
                 & Agricultural                                                          &                      \\ \cline{2-3} 
                 & Urban                                                                 &                      \\ \cline{2-3} 
                 & Industrial                                                            &                      \\ \cline{2-3} 
                 & Grassland                                                             &                      \\ \cline{2-3} 
                 & Desert                                                                & x                    \\ \cline{2-3} 
                 & Sea                                                                   & x                    \\ \hline
Fire   size      & Small \textless{}30x30m                                               & x                    \\ \cline{2-3} 
                 & 30x30m\textless{}=Small-medium\textless{}60x60m                       &                      \\ \cline{2-3} 
                 & 60x60m\textless{}=Medium-large\textless{}90x90m                       &                      \\ \cline{2-3} 
                 & Large \textgreater{}=90x90m                                           &                      \\ \hline
Fire   intensity & Low \textless Schroeder conditions1                                   & x                    \\ \cline{2-3} 
                 & Medium \textgreater Schroeder conditions                              &                      \\ \cline{2-3} 
                 & High \textgreater{}\textgreater Schroeder conditions1                 &                      \\ \hline
Clouds           & None                                                                  &                      \\ \cline{2-3} 
                 & Low coverage\textless{}25\% of tile                                   &                      \\ \cline{2-3} 
                 & 25\% of tile\textless{}=Low-medium   coverage\textless{}50\% of tile  &                      \\ \cline{2-3} 
                 & 50\% of tile\textless{}=Medium-high   coverage\textless{}80\% of tile &                      \\ \cline{2-3} 
                 & High coverage \textgreater{}80\% of tile                              &                      \\ \hline
Time of   day    & Early morning 7-9 am                                                  &                      \\ \cline{2-3} 
                 & midday 12-14                                                          &                      \\ \cline{2-3} 
                 & late afternoon 4-6                                                    &                      \\ \cline{2-3} 
                 & Night                                                                 & x                    \\ \hline
Time of   Year   & Winter                                                                &                      \\ \cline{2-3} 
                 & Spring                                                                &                      \\ \cline{2-3} 
                 & Summer                                                                &                      \\ \cline{2-3} 
                 & Autumn                                                                &                      \\ \hline
\end{tabular}
\end{center}
\end{table}

\subsubsection{Rationale for ML Safety Requirements}

In this section the rationale for how each of the ML safety requirements was derived is provided. Note that there were no ML safety requirements specified relating to system safety requirement REQ-SAFE-ER-2 since this requirement relates to the revisit rate of the satellite and the communication time of the generated fire alert to the emergency services. As such the ML component does not contribute to the satisfaction of this requirement.

\textbf{MLSR1} - This requirement is derived from system safety requirement REQ-SAFE-ER-1. For the images used on these type of CubeSat satellites, 6 pixels represents 180m, so this requirement will ensure that the actual fire is never more than 180m from a reported position.

\textbf{MLSR2} - This requirement is derived from system safety requirement REQ-SAFE-ER-3. The current standard for image-based fire detection is that provided by the Fire Information for Resource Management System (FIRMS)\footnote{https://earthdata.nasa.gov/earth-observation-data/near-real-time/firms}\footnote{FIRMS provides active fire data from NASA’s Moderate Resolution Imaging Spectroradiometer (MODIS) and Visible Infrared Imaging Radiometer Suite (VIIRS) instruments}. FIRMS achieves an omission error rate of ~5\% \cite{giglio2016collection}, which the on-board fire alert system must match. The Schroeder conditions represent an accepted threshold for labelling of active fires in satellite data \cite{schroeder2016active}.

\textbf{MLSR3} - This requirement is derived from system safety requirement REQ-SAFE-ER-4. The key consideration for this requirement was that false alerts shouldn't happen so frequently that they become hazardous. This could happen either through diverting fire response resource to a region of no fire and away from areas where the fire response is required. Or it could become hazardous through becoming a nuisance to operators who then start to ignore genuine alerts or even turning the system off. It should be noted that the fire alerts provided by the satellite would not be the only source of information available to responders, who may have the opportunity to corroborate with more local ground-based fire observation. Again FIRMS was taken as the current standard for false positive performance in fire detection. In an average month, FIRMS detects around 5,000 wildfires in the US, approximately 52 of which, on average, are false positives.

\textbf{MLSR4} The performance of fire detection algorithms can vary substantially depending on a number of key factors \cite{schroeder2016active}. Table \ref{tab:Rob} captures features of the image data that represent the variation in these factors that must be considered in the data sets in order to provide coverage of the operating domain of the system. 

\subsubsection{ML Requirements Assurance Argument}

\begin{figure}[h]
    \centering
    \includegraphics[width=1\linewidth]{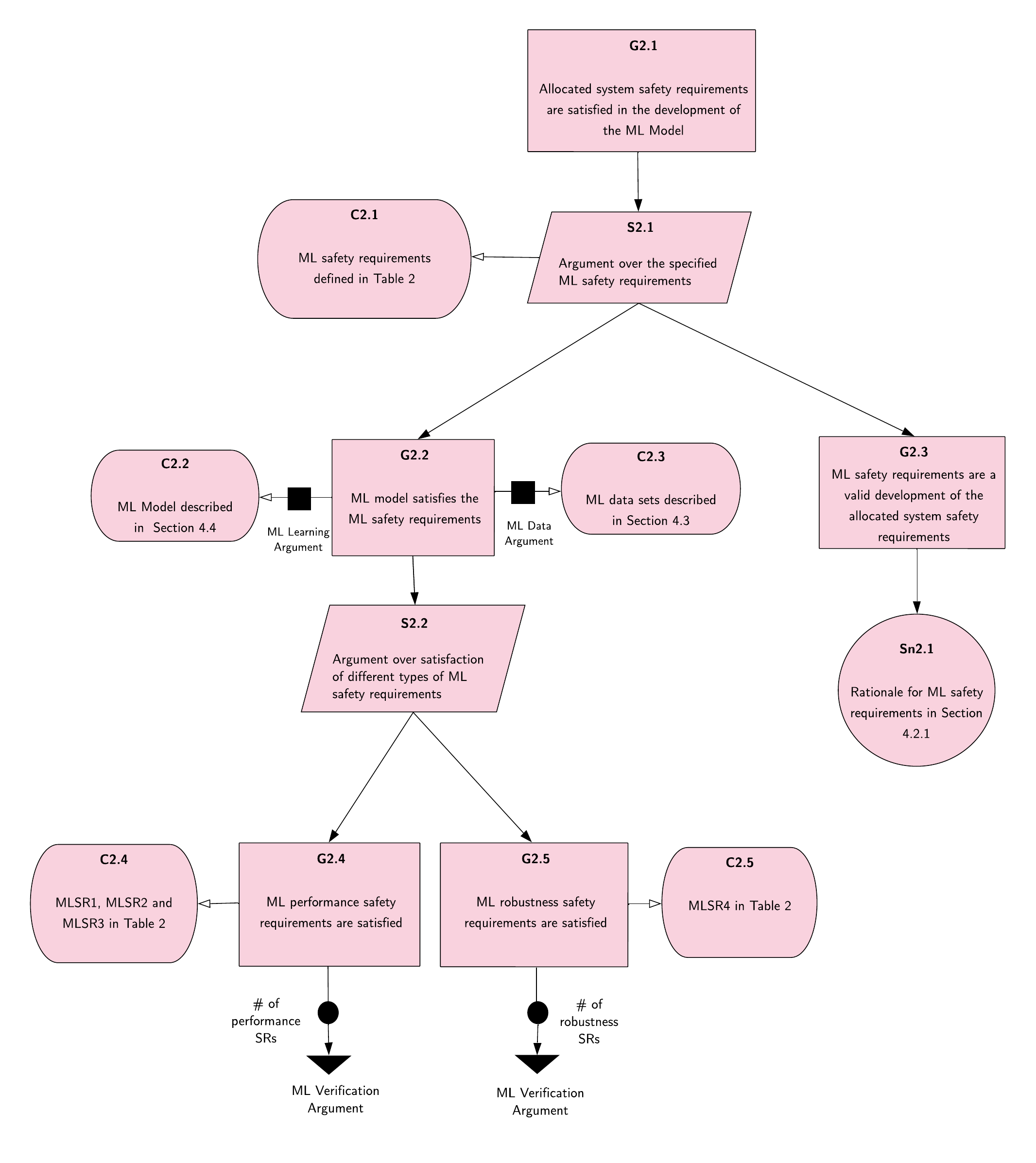}
    \caption{ML Safety Requirements Assurance Argument for wildfire alert component}
    \label{fig:MLRArg}
\end{figure}

Figure \ref{fig:MLRArg} shows the part of the ML component argument relating to the ML safety requirements. The argument splits into two safety claims: 
\begin{itemize}
    \item A claim that the ML safety requirements are correctly defined (G2.3). This is supported by evidence regarding the rationale for the requirement definition (Sn2.1).
    \item A claim that the ML model satisfies the defined ML requirements (G2.2). Here the argument is split to separately consider the performance and robustness requirements. For each of these safety claims, verification will be used to generate evidence to demonstrate that the ML safety requirements are satisfied. This is discussed further when describing the verification argument in section \ref{sec:verif}.
\end{itemize}

The ML requirements satisfaction claim (G2.2) can be seen to be presented in the context of the ML model and the ML data. Arguments regarding the sufficiency of the data and the learned model have been developed, and are presented in sections \ref{sec:model} and \ref{sec:data}. These argument connect to figure \ref{fig:MLRArg} at the assurance claim points (ACPs) indicated by the black squares.

\subsection{Data Management Assurance}\label{sec:data}

Data plays a particularly important role in machine learning since data encodes the requirements which
will be embodied in the resulting ML model. It is therefore crucial as part of the safety case for the ML component to demonstrate that the data is sufficient to ensure that the learned model will satisfy the ML safety requirements. At this stage we therefore carried out the following activities:
\begin{enumerate}
    \item Defined data requirements against which the data sets produced could be assessed.
    \item Generated data sets that satisfied the specified data requirements.
\end{enumerate}

\subsubsection{Data Requirements}

The ML data requirements relating to the wildfire detection ML component are described below. ML data requirements have been specified for relevance, completeness, accuracy and balance of the data. Requirements relating to relevance specify the extent to which the data must match the intended operating domain into which the model is to be deployed. Requirements relating to completeness specify the extent to which the data must be complete with respect to a set of measurable dimensions of the operating domain. This is done by considering the dimensions of variation that were identified in Table \ref{tab:Rob} as part of the ML safety requirements. Requirements relating to accuracy specify how the accuracy of the information in the data sets will be judged. Requirements relating to balance specify the required distribution of samples in the data sets. A balanced data set is one with an appropriate number of samples for each class or feature of interest. Note that this does not necessarily mean that an equal number of samples is required for each class; rare classes may require fewer samples in order to be balanced. Table \ref{tab:DSRs} presents the ML data requirements specified for each of these properties for the wildfire alert ML component.

\begin{table}[]
\caption{\label{tab:DSRs}ML data requirements for wildfire alert system ML component}
\begin{tabular}{@{}ll@{}}
\toprule
\multicolumn{2}{l}{\textbf{Relevance}}                                                                                                                                                                                                                                                                  \\ \midrule
\multicolumn{1}{l|}{\textbf{DR1}} & Only data samples of areas of the specified land type shall be included in the data sets.                                                                                                                                                                           \\ \midrule
\multicolumn{1}{l|}{\textbf{DR2}} & \begin{tabular}[c]{@{}l@{}}The format of each data sample shall be representative of images captured using sensors deployed \\ on the target satellite. This shall include a representative resolution, spectral band and image size.\end{tabular}                  \\ \midrule
\multicolumn{1}{l|}{\textbf{DR3}} & \begin{tabular}[c]{@{}l@{}}Each data sample shall represent a sensor position which is representative of that to be used on the\\ target satellite. This shall include consideration of the angle, height and field of view of the deployed \\ sensor.\end{tabular} \\ \midrule
\multicolumn{2}{l}{\textbf{Completeness}}                                                                                                                                                                                                                                                               \\ \midrule
\multicolumn{1}{l|}{\textbf{DR4}} & \begin{tabular}[c]{@{}l@{}}The data sets shall include samples representing combinations of each of the in-context element \\ classes defined in Table \ref{tab:Rob}.\end{tabular}                                                                                              \\ \midrule
\multicolumn{1}{l|}{\textbf{DR5}} & The data sets shall include samples containing fires and no fires.                                                                                                                                                                                                  \\ \midrule
\multicolumn{2}{l}{\textbf{Accuracy}}                                                                                                                                                                                                                                                                   \\ \midrule
\multicolumn{1}{l|}{\textbf{DR6}} & All masks generated shall be sufficiently large to include the entirety of the fire                                                                                                                                                                                 \\ \midrule
\multicolumn{1}{l|}{\textbf{DR7}} & \begin{tabular}[c]{@{}l@{}}All masks generated shall be no more than 6 pixels larger in any dimension than the minimum \\ sized mask capable of including the entirety of the fire\end{tabular}                                                                     \\ \midrule
\multicolumn{1}{l|}{\textbf{DR8}} & All data sample with fires present in the data samples must be correctly labelled                                                                                                                                                                                   \\ \midrule
\multicolumn{1}{l|}{\textbf{DR9}} & \begin{tabular}[c]{@{}l@{}}The labels for the position of fires within each image must be no more than 6 pixels outside the \\ boundary of the area of the real fire.\end{tabular}                                                                                                                             \\ \midrule
\multicolumn{2}{l}{\textbf{Balance}}                                                                                                                                                                                                                                                                    \\ \midrule
\multicolumn{1}{l|}{\textbf{DR10}} & \begin{tabular}[c]{@{}l@{}}The data sets shall include a suitable distribution of samples for each combination of element classes \\ defined in Table 1 of ML safety requirements document.\end{tabular} 
                                                   \\ \bottomrule
\end{tabular}
\end{table}

\textbf{\emph{Rationale for ML Data requirements}}

Here we describe the rationale for each of the data requirements.

\textbf{DSR1} - The wildfire alert system is not expected to operate over all areas. Images that represent areas out of the defined intended scope of operation should not be included in the data sets.

\textbf{DSR2} - The satellite will provide images to the ML component with a particular format. Therefore only images of that format should be used in the development of the model.

\textbf{DSR3} - The satellite will provide images to the ML component that are taken from a particular position and orientation. Therefore only images that exhibit equivalent characteristics should be used in the development of the model.

\textbf{DSR4} - The operating domain of the satellite is defined by the features in the table. We must ensure that the data sets include data items for each combination of these features.

\textbf{DSR5} - The satellite needs to avoid false positives so the data sets must include examples of images without a fire present.

\textbf{DSR6} - The mask must be big enough that none of the fire is missed.

\textbf{DSR7} - The mask must not be so big that any positions identified by the mask are too far from the actual position of the fire.

\textbf{DSR8} - If fires are present but not labelled then the data will be incorrect.

\textbf{DSR9} - The data must be labelled with sufficient accuracy, see rationale for MLSR1

\textbf{DSR10} - No element class should be under or over represented as this will result in inconsistent and biased performance. The number of data items required of each class may not be equal. The distribution across the classes in each data set should be justified as part of data management.

\subsubsection{Data Generation}

Three separate datasets were created development data, internal test data and verification data. The
first two of these sets are for use as part of the development of the model (see section \ref{sec:model}). The verification set is used in model verification. The focus of this data set is therefore not on creating a model (as for the other two sets) but instead on finding realistic ways in which the model may fail when used in an operational system. It is crucial therefore that the verification data is generated independently from the development process. The verification data is discussed in more detail in section \ref{sec:verif}.

The development and internal testing data was generated from the large Landsat-8 data set \cite{t9gn-y009-20}. This was felt to be an appropriate source of data for this application for a number of reasons. Truth masks are available for the data which enables pixel level classification of active fire. The truth masks are arrays, which allows for configuration of image tile size. The dataset is large in size and contains imagery covering all of South America with a variety of land types and land uses. It provides coverage of various fire sizes, distributions and intensities. The imagery contains 10 spectral bands of data for each capture. The Landsat-8 sensor has 30 metres of spatial resolution, meaning one pixel is equivalent to 30m\textsuperscript{2} in ground area. Labels on the image data are created using a complex set of conditions, based on information contained in 7 bands of the satellite data, plus associated meta-data. There were also some limitations to this data that had to also be considered. Firstly, the data set contains images captured in the year 2018 only and covers only South America. Also, the labels on the data have not been manually corrected and will therefore be expected to include a small level of error. In particular, instances of intense heat in urban settings may be falsely labelled as active fire.

Some pre-processing was carried out on the data before creating the data sets. Firstly, of the 10 spectral bands available 3 were chosen: Blue, SWI-1 and SWI-2. This combination, including both short wave infrared channels, has previously been shown to be successful for creating models for active fire detection \cite{de2021active}. Secondly, the dataset contained image tiles of 128 x 128 pixels. The learned model needs to perform on continuous data on the satellite which is cropped into tiles of 48 x 48 pixels. The selected image data was therefore cropped from 128 x 128 to 48 x 48 image tiles. 

Two sets of internal test data were created. Set 1 is a subset of the same dataset from which the development data was generated (\cite{t9gn-y009-20}). Set 2 is a collection of unlabelled data captured by Landsat-8 over the US state of Oregon (a target area of interest for the application), downloaded via Sentinel Hub\footnote{https://www.sentinel-hub.com/}. Set 2 was used to carry out initial, internal testing of the model performance on data from the area of interest, and to introduce some edge cases. 

\subsubsection{Data Evaluation}

The development and internal testing data sets were evaluated against the defined data requirements (table \ref{tab:DSRs}). Below we summarise the results of the data evaluation.

\textbf{\emph{Relevance:}} A subset of the Landsat-8 data was selected covering areas of South America including Chile and Argentina as well as Oregon. These areas were chosen in particular since they contain large areas of temperate rainforest ensuring images relevant to the application domain are provided. The size and spectral range of the images is equivalent to the operational images generated on-board the satellite.    

\textbf{\emph{Completeness:}} As well as providing large areas of temperate rainforest, the chosen regions is are sufficiently geographically diverse to provide image samples of urban land, agricultural and grazing land. While the data was captured across a single year, it has a temporal resolution of 16 days and so contains samples taken at the same locations at different times throughout the year.Various cloud level samples were gathered for both non fire and fire instances. Samples containing large reflective surfaces were included to test for false positive detection. Samples containing fire of various size and spread were gathered.

\textbf{\emph{Accuracy:}}

The labelling conditions used to generate the truth masks in the Landsat-8 data set are complex and well documented \cite{de2021active}. To provide validation for the truth masks, visual comparisons were made between the truth masks and the images viewed in the visual range. While a small level of error was seen within the subset, this is a common and acceptable limitation of large, labelled datasets. 

\textbf{\emph{Balance:}} 

A review showed that a good balance of the various features and locations was achieved across the data sets. There are far greater instances of non-fire pixels than of fire pixels in the available images. The development data sets therefore included more images featuring some fire pixels to ensure better balance.

\subsubsection{ML Data Assurance Argument}

\begin{figure}[h]
    \centering
    \includegraphics[width=0.8\linewidth]{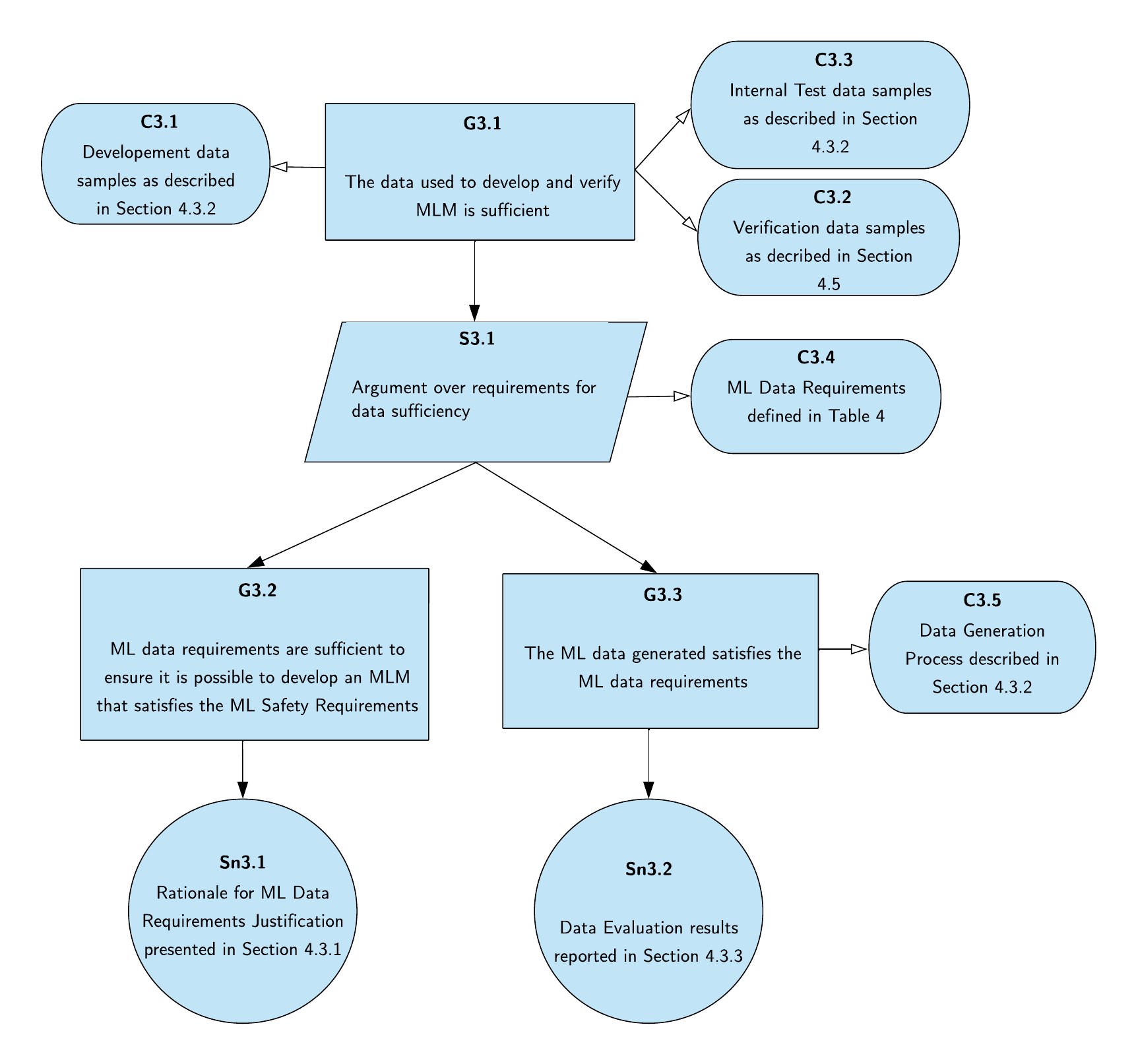}
    \caption{ML Data Assurance Argument for wildfire alert component}
    \label{fig:MLdataArg}
\end{figure}

Figure \ref{fig:MLdataArg} shows the part of the ML component argument relating to the data.
The argument presents a claim that the data used to develop the ML model is sufficient from a safety assurance perspective (G3.1). The context for this claim is the three datasets that were generated. The argument to support the claim considers the data requirements. Two claims are made. Firstly, that those data requirements are good enough to ensure that the ML safety requirements are satisfied (G3.2); this is demonstrated using the documented rationale for the data requirements (Sn3.1). Secondly, that the specified data requirements are satisfied by the generated data (G3.3); this is demonstrated through the results of the data evaluation (Sn3.2).

\subsection{Model Learning Assurance}\label{sec:model}

At this stage of the process the development data created at the previous stage was used to create candidate models that were able to satisfy the defined ML safety requirements. The candidate models that were created were tested using the internal test data in order to select the best model to use. 

\subsubsection{Model creation}

Tensorflow\footnote{https://www.tensorflow.org/} was selected as the tool for developing the wildfire alert model, since it is a well-established and well documented tool. Tensorflow also comes with a visualisation tool, Tensorboard \footnote{https://www.tensorflow.org/tensorboard}, which enables monitoring of different metrics during the training process and allows easy comparison of differences between training runs with alternative parameter settings.

The Unet architecture was used as it is a popular CNN model for pixel classification (semantic segmentation) which has been shown to be successful in performing active fire detection on a large dataset \cite{de2021active}. The network consists of a contracting path and an expansive path, which gives it a u-shaped architecture.  Two variations of Unet were developed: Unet-128 and Unet-48. Initially the Unet 128 was selected for training using data of size 128 x 128. It was found during development however that significant pixel areas of active fire were classified as false negative by the model. Data processing was therefore carried out to split the 128 x 128 images into 48x48 samples, to address the lack of ‘clipped’ fire areas in the labelled data and make it more representative of the kind of real-world data the model will be applied to. To adapt the model to work well with 48 x 48 input, the layer values throughout the model were reduced incrementally to find the optimal combination. 

Binary Cross Entropy Loss is a popular and successful loss function for binary classification problems. The predicted class probability is compared to the actual class, and the resulting score considers how far apart these values are. The Dice Coefficient represents the size of the overlap of the segmentation class in each mask, divided by the total size of the two images. The sum of the Binary Cross Entropy Loss and Dice Coefficient Loss was used as a custom loss function during training and was found to gain better results than either metric used alone, or alternative loss metrics.

Both Stochastic Gradient Descent (SGD) and Adam were used as methods for optimising the objective function during model learning. Adam differs from SGD in that the learning rate is not static throughout training. With the Adam optimiser, a learning rate is maintained for each model parameter and adapted as training progresses. It is an easily configurable optimiser, where the default parameters perform well on most problems \cite{kingma2014adam}.  It is a popular optimiser for deep learning with large datasets, as good results can be reached quickly, and it was found to achieve the best performance during development of the wildfire alert model.

The learning rate for the model was initialised at 0.1 and incrementally decreased to find the best performing value. A learning rate of 0.01 was found to yield the best performance during training.

\subsubsection{Internal testing approach}

The performance of the candidate models created was measured using the Mean Intersection over Union (Mean IoU) value between the label mask and the model output mask.  Intersection over Union (IoU) is the area of overlap, divided by the area of union between the label and output masks. The metric ranges from 0 to 1 with 0 signifying no overlap and 1 signifying perfectly overlap-ping masks. The Mean IoU for the classification is calculated by taking the IoU of each class (fire and non-fire) and averaging them.  Mean IoU is a very useful metric for semantic segmentation problems where there is class imbalance, providing a much more meaningful representation of how well the model output mask matched the truth mask than a simple pixel accuracy score.

\subsubsection{Internal test results}

\begin{figure}[h]
    \centering
    \includegraphics[width=0.6\linewidth]{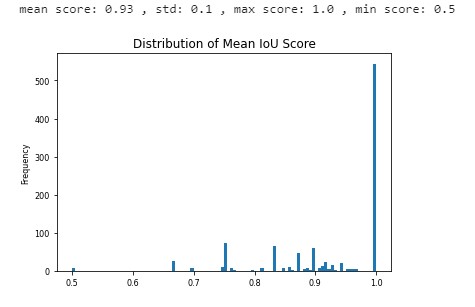}
    \caption{Mean IoU score frequency across internal data set 1}
    \label{fig:internal1}
\end{figure}

\begin{figure}[h]
    \centering
    \includegraphics[width=1\linewidth]{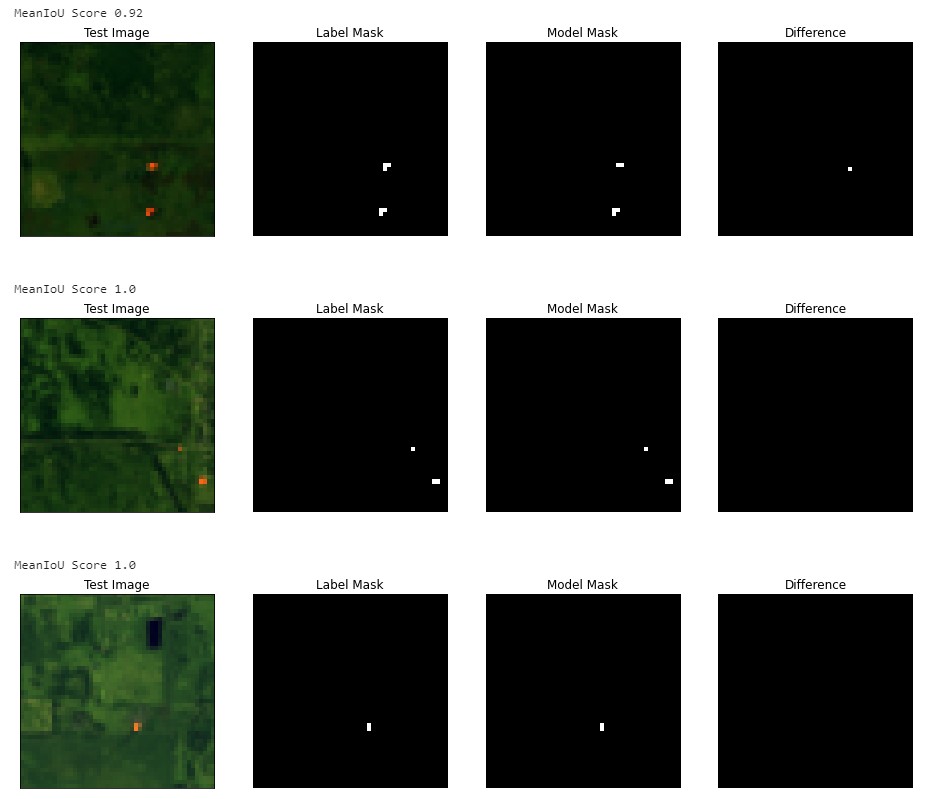}
    \caption{Randomly sampled images from internal data set 1}
    \label{fig:internal1ex}
\end{figure}

Two internal test data sets were used. The first was a set of 1000 image tiles with corresponding truth masks. The Mean IoU scores of the fire class and the non-fire class were calculated for the entire set. The average Mean IoU score on the test set was 0.93. Figure \ref{fig:internal1} visualises the distribution of scores across the set as a whole. Figure \ref{fig:internal1ex} shows a comparison of the model output mask and the truth mask for randomly selected sample of images from the data set along with a visual comparison of the difference. The Mean IoU score for each of these samples is displayed.

To asses the performance of the models against the ML safety requirements it was necessary to use the IoU scores to quantify the false positives and the false negative detections of active fires in the data samples. A threshold on the IoU scores for both the active fire and non fire class was used to generate false positive and false negative values for the model performance. 
The values were calculated as follows:

\begin{itemize}
    \item False Negative: model mask and truth mask have an IoU score below threshold (calculated for active fire class).
    \item False Positive: model mask and truth mask have an IoU score below threshold (calculated for non fire class).
\end{itemize}

The following threshold values were selected by analysing the IoU scores for each class to define meaningful false positive and false negative values:

\begin{itemize}
    \item False Negatives: where the IoU score for the fire class is less than 0.3
    \item False Positives: where the IoU score for the non-fire class is less than 0.99
\end{itemize}

For internal test set 1 (containing 1000 samples), 0 false positives were found, and 8 false negatives were found, which translates to 0.8\% of the set. 

A second set of internal test data was generated to verify the model performance against continuous data. Continuous data is also more relevant to the way the model will be executed in operation. This was done by downloading a selection of large images of size 2000 x 1600 pixels. The images were split into 1428 tiles, of size 48 x 48 pixels, suitable for the model.  The model produced 1428 output masks which were assembled to create a large output mask. A visual comparison was then made between the large image and the large model output mask with no false negatives identified.

The results obtained from internal testing were compared to the defined ML saftey requirements in order to assess the sufficiency of the model. Below we discuss each of the ML safety requirements in turn.

\begin{itemize}
    \item MLSR1 - From analysis of IoU and Mean IoU scores between the model output masks and the truth masks, the model is therefore seen to satisfy the requirement since the recorded error was always less than 6 pixels in any direction when executing the model against the internal test data.
    \item MLSR2 - Across the internal test data, a false negative rate of 0.8\% was found. The model is therefore seen to satisfy the requirement as it positively identified 99.2\% of all visible active fires across the test data.
    \item MLSR3 - The model was seen not to make any false positive detections across the internal test data. 
\end{itemize}

\subsubsection{ML Learning Assurance Argument}

\begin{figure}[ht]
    \centering
    \includegraphics[width=0.7\linewidth]{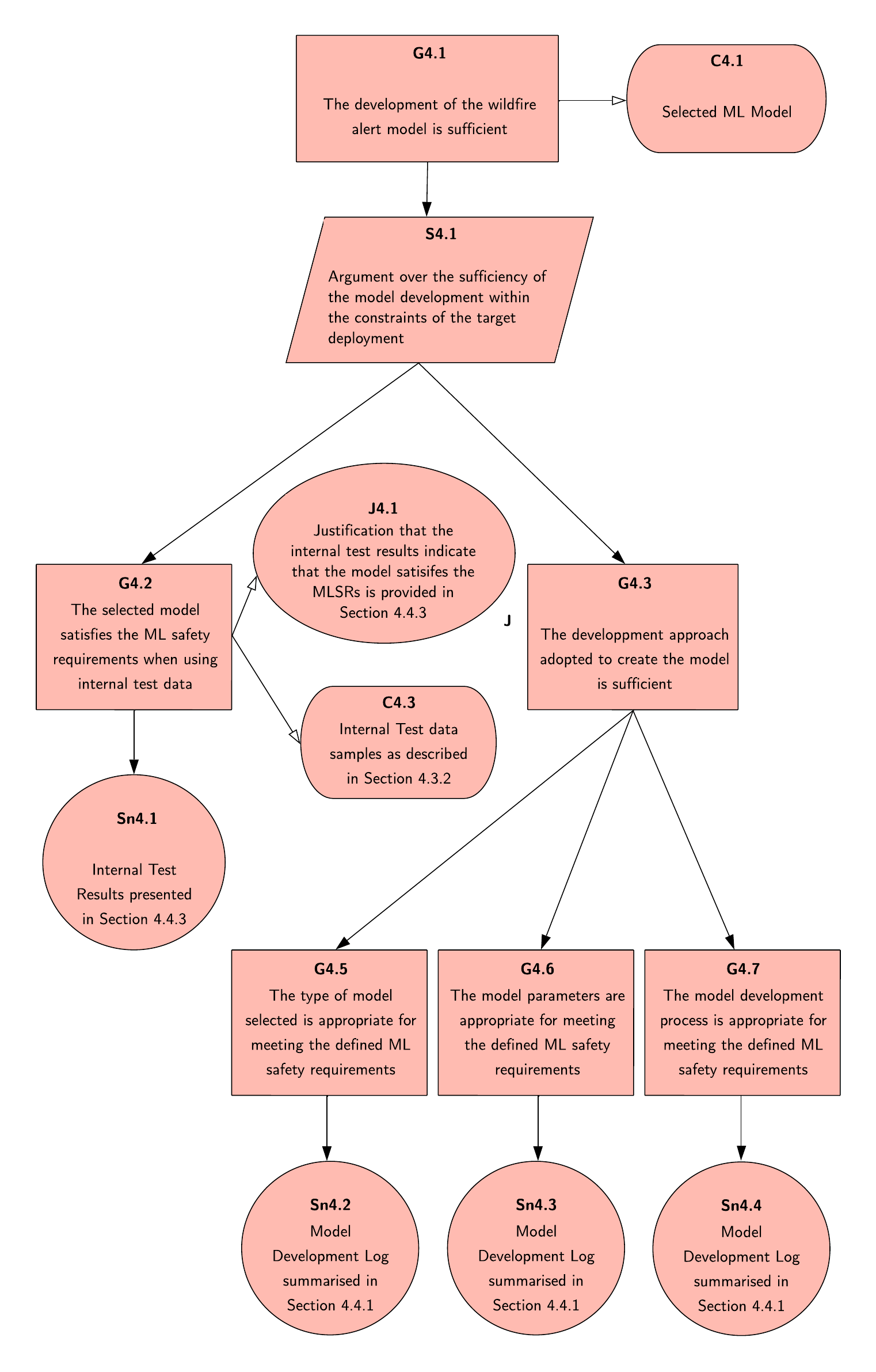}
    \caption{ML Learning Assurance Argument for wildfire alert component}
    \label{fig:LearnArg}
\end{figure}

Figure \ref{fig:LearnArg} shows the part of the ML component argument relating to the model learning.
The argument presents a claim that the way in which the ML model was developed is sufficient given the constraints that are imposed by the platform to which the model is being deployed (G4.1). An argument is made to support this by showing that the selected model satisfies the defined ML safety requirements (G4.2). The results that are observed from executing the model with the internal test data are used as evidence for this, and a justification is also provided as to how the observed results indicate that the ML safety requirements are satisfied (J4.1). In addition a claim is made that the development approach itself that was used to create the model is sufficient (G4.3) This claim is supported by consideration of the type of model used model parameters, as well as the nature of the learning process itself that was adopted. All of these ML development decisions were recorded and justified in a model development log.

\subsection{Model Verification Assurance}\label{sec:verif}

\subsubsection{Verification Data}

Verification data was collected by a team of people who were not involved in the development of the ML model. The verification data provided images for the ML model with the following characteristics:

\begin{itemize}
    \item Representative of images that may be observed by the satellite during operation in locations and conditions within the scope defined for the safety case of the ML component in section \ref{sec:scope}
    \item Provide examples across the range of the key features relevant to the use of the ML component  as identified in Table \ref{tab:Rob}
    \item Represent interesting or challenging cases within the scope of use (i.e. edge cases)
\end{itemize}

With these criteria in mind, below we discuss the key features considered in generating the verification data.

\textbf{\emph{Land Type}}

 The appearance of fire may be different if the fire occurs on different types of terrain. To check that the model is generalisable, a range of images of the different land types were included in the verification data. Image samples of each land type captured by the Landsat-8 satellite were downloaded via Sentinel Hub. The areas from which images were chosen to represent each land type were:	
 
 \begin{itemize}
    \item Temperate rainforest - New Zealand, where all areas are classed as temperate rainforest.
    \item Agricultural - North Dakota, where 90\% of the land that makes up North Dakota is used for farms and ranches. 
    \item Urban - Greater Tokyo Area, which is the most populous metropolitan area in the world 
    \item Industrial - Southern New England, which has extensive areas of diversified industrial growth
    \item Grassland - Canada Prairie, where large areas of Alberta, Saskatchewan, and Manitoba are temperate grassland and shrubland.
\end{itemize}

By referring to information on the locations of active fires in the FIRMS database it was possible to download images within each of these geographical regions that were known to contain wildfires.

\textbf{\emph{Fire Size}}

 In order to check whether the size of the fire affected the performance of the learned model, images with fires of different sizes within each of the chosen regions were selected. For the purposes of verification data we selected images that had either small (<30m longest dimension) or large (>100m longest dimension) fires. In addition, images were included in the verification data set that did not contain active fires. This was to provide verification of the false-positive performance of the ML component. The development team were not aware which of the images in the verification data set contained fires.
 
\textbf{\emph{Cloud Cover}}

In order to check whether the presence of cloud cover in the image affected model performance, images containing different levels of cloud were selected. Images with no clouds, with low cloud cover (<10\% of image) and high cloud cover (>50\% of image) were selected.

\textbf{Verification Test Cases}

The images used as verification test cases were chosen by considering combinations of the features discussed above in order to provide sufficient coverage. Where relevant, in each case the specific images chosen were assessed as containing interesting or unusual features. Figure \ref{fig:verifResults} identifies each of the cases for which a verification image was obtained.

\subsubsection{Verification Results}

\begin{figure}[h]
    \centering
    \includegraphics[width=0.7\linewidth]{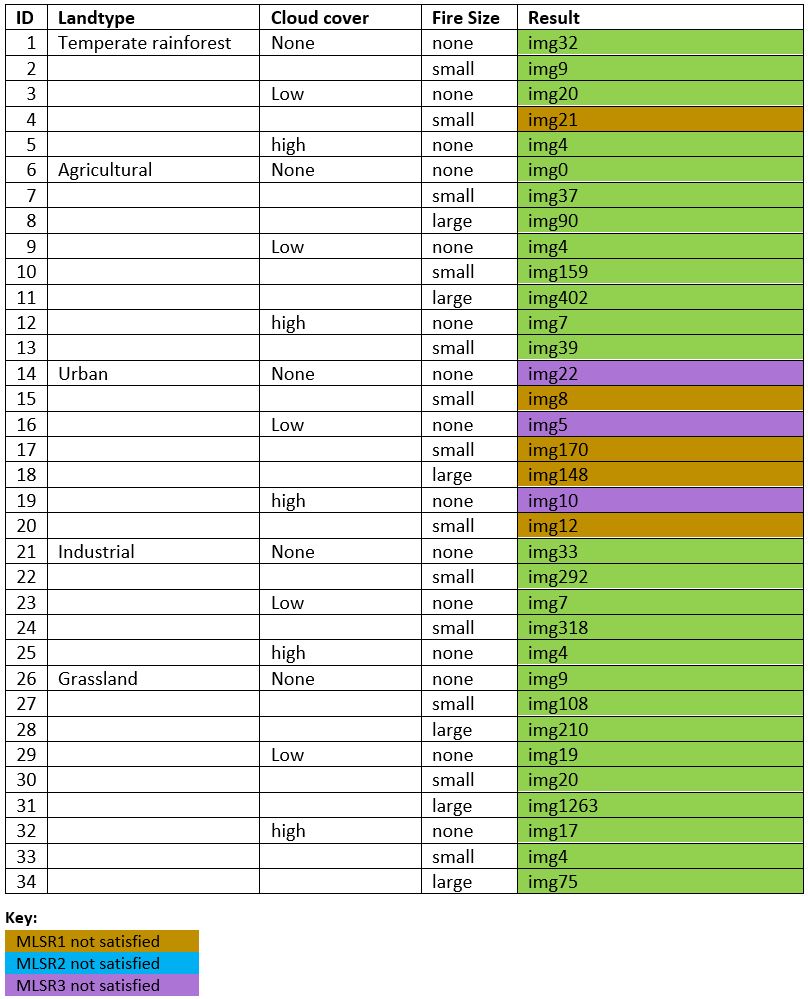}
    \caption{Verification results obtained for wildfire alert model}
    \label{fig:verifResults}
\end{figure}

The results are presented in Figure \ref{fig:verifResults} for each of the verification images. The results column shows colours to indicate the result. Green indicates that all the MSRs were satisfied for that image. The other colours indicate that one of the MSRs was not satisfied as defined in the key.

\begin{figure}[h]
    \centering
    \includegraphics[width=0.8\linewidth]{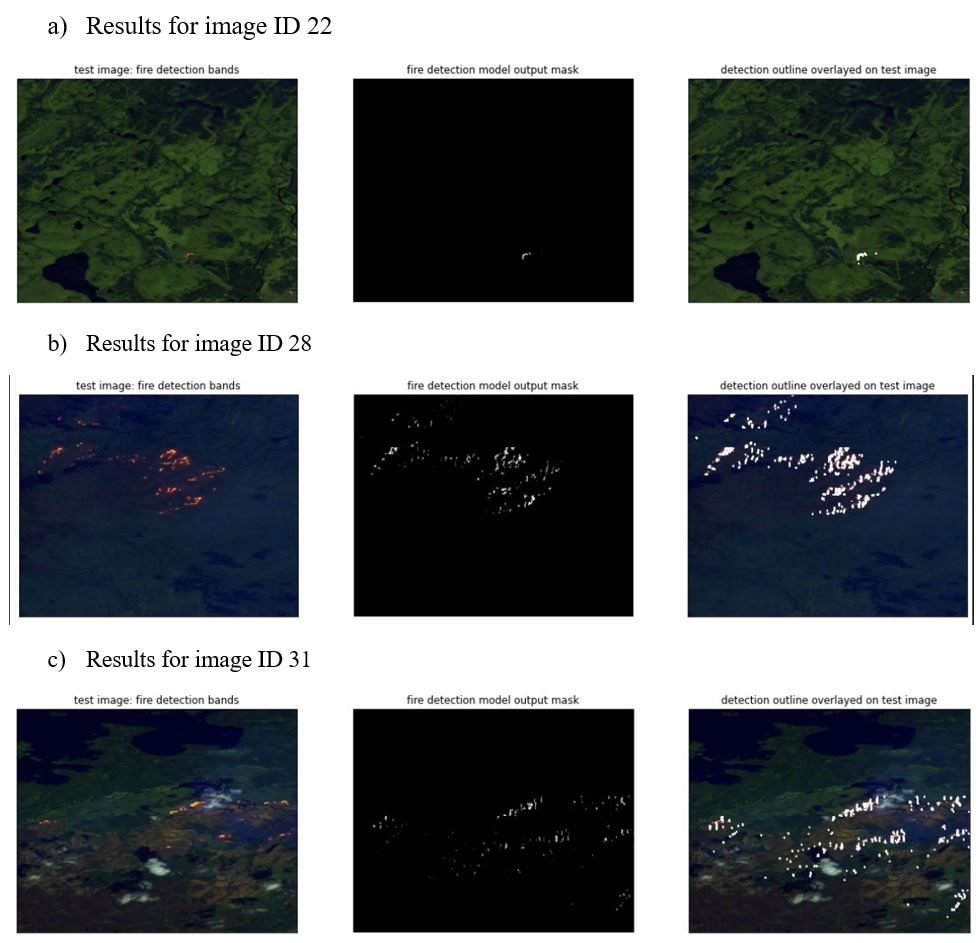}
    \caption{Example verification outputs}
    \label{fig:verifImage}
\end{figure}

Examples of the outputs for three of the verification images are shown in figure \ref{fig:verifImage}. These show, for each case, the test image, the output mask generated by the ML component, and the mask overlayed over the image.

\textbf{Verification Findings}

It can be seen from the results presented in Figure \ref{fig:verifResults} that none of the verification images obtained from an urban area satisfied the MSRs. In all cases there were a large number of false detections observed in the output. These results indicate that the model is not suitable for detecting active fires in urban areas and this should be explicitly documented as a limitation of use within the safety case.

Of the remaining cases there was just one image that didn’t satisfy the MSRs. This was case ID 4 where the position of the output mask was not sufficiently aligned with the true fire position. The reasons for this anomaly are unclear and are the subject of further investigation.

\subsubsection{Verification Argument}

\begin{figure}[h]
    \centering
    \includegraphics[width=0.8\linewidth]{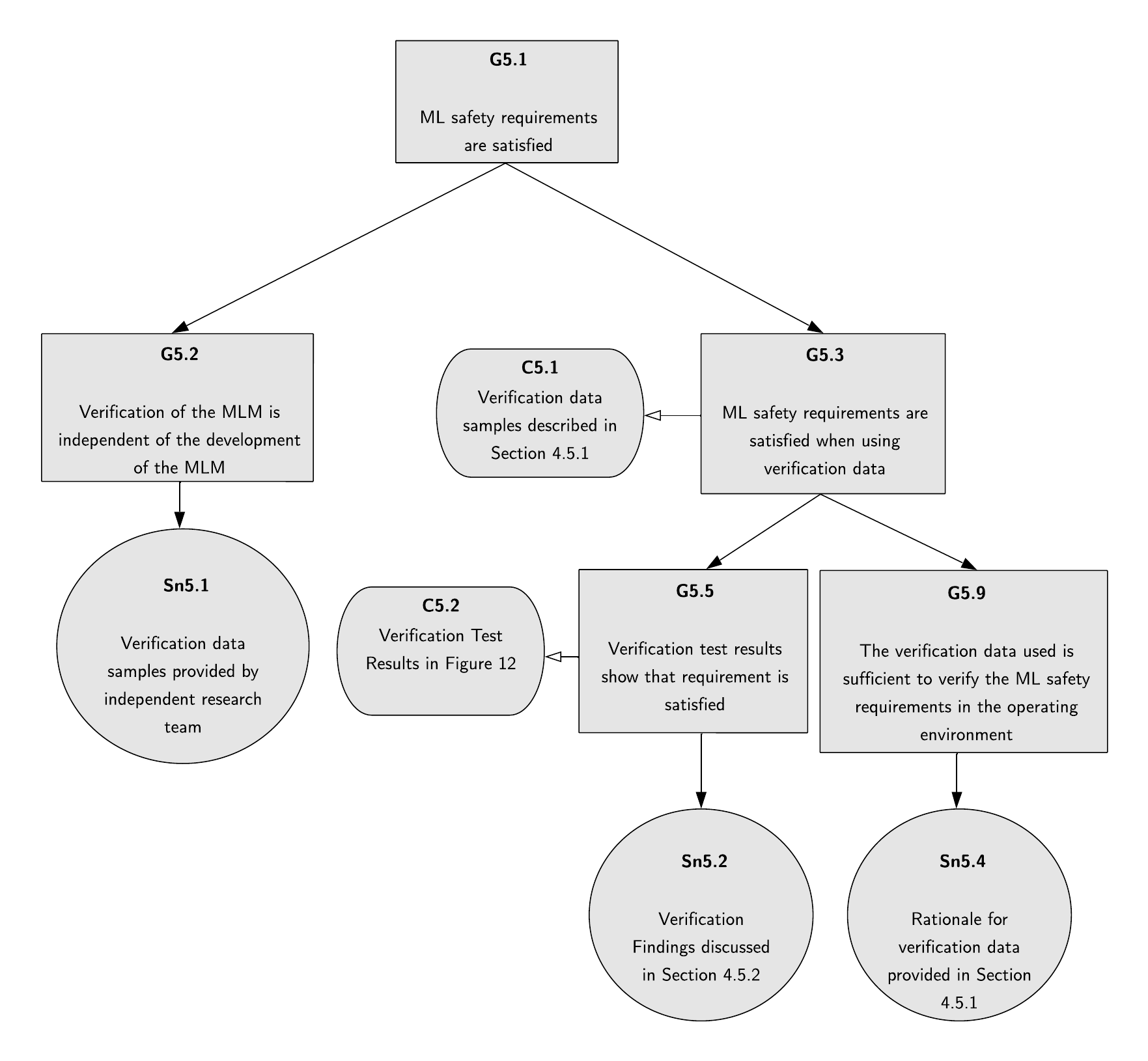}
    \caption{ML Verification Argument for wildfire alert component}
    \label{fig:verifArg}
\end{figure}

Figure \ref{fig:verifArg} shows the part of the ML component argument relating to the model verification. There are two main claims that are made as part of the verification argument. Firstly it is demonstrated that the verification of the ML model is independent from the development of the model (G5.2). In this case it can be shown that the verification data used was collected by a team from another organisation that did not develop the ML model. Secondly, a claim is presented that when this data is provided to the ML model, the ML safety requirements are satisfied (G5.3). This claim is supported by providing the verification test results themselves, along with an explanation as to how those results show satisfaction of the safety requirements. This also requires that the sufficiency of the verification data that was used is demonstrated (G5.9).

\subsection{Model Deployment Assurance}\label{sec:deploy}

The aim of this stage of the process is to demonstrate that the system safety requirements for which the model has been developed continue to be satisfied when the model is integrated into the overall satellite system and operates in the real environment. Since the wildfire alert component has not yet been deployed to the satellite, this stage of the process has been limited to integration testing using hardware-in-the-loop (HIL) simulation to recreate, as closely as possible the deployment environment for the wildfire alert component. The simulation employed real multi-spectral optical data captured over the deployment region (Oregon) by the Landsat-8 satellite and sourced from Sentinel Hub. 

The simulations were performed using a number of different operational scenarios representing satellite passes over Oregon at locations and times with different numbers of visible active fires. It is expected that the wildfire alert component detects the fires and generates an alert indicating the size and locations of the fires.



\subsubsection{Integration test results}

It was necessary to assess whether the integration tests indicate that the system safety requirements were satisfied. The preferred strategy was to undertake a comparison with NASA FIRMS fire detections for the same date and location to validate the geolocation accuracy of the processing chain. However, the FIRMS detections were determined not to be a reliable ground truth because there is a significant time difference between the capture made by the VIIRS sensor from which the FIRMS detections were made, and the capture made by the Landsat sensor which has been used as test data. Instead, a visual inspection of the detections was made. The masks were analysed along with the fire detection bands of the data. When these fire detection bands are displayed, active fire pixels appear as a bright blue colour. During analysis, ambiguous cases were found. Three different approaches were taken in order to try to eliminate subjectivity of such cases when defining false negatives and false positives. 

\begin{itemize}
    \item Approach 1: Generous 
    \begin{itemize}
        \item Pixels of a darker and/or duller blue which have not been classified as containing active fire, are considered to be true negatives. 
        \item Areas where the mask has union with but does not cover the entirety of visible active fire, are considered to be complete detections. Pixels not covered by the mask in these cases are not considered to be false negatives. \item Pixels of a bright or middle shade blue that are small in area and distant from other detections or ambiguous/non-active fires are considered to be true positives. 
    \end{itemize}
    
    \item Approach 2: Moderate 
    \begin{itemize}
        \item Pixels of a darker and/or duller blue which have not been classified as containing active fire, may be considered to be false negatives. This distinction depends mainly on the brightness of the blue colour.  
        \item Areas of pixels of a middle shade blue colour are counted as discrete active fires if they are distant or moderately close to another detection, or another ambiguous fire.
        \item Pixels of a darker and/or duller blue may be considered a false detection if area is small and distant from other detections. Small pixel areas that are bright blue are considered as false positives if the general location appears to be built up. 
    \end{itemize}
    
    \item Approach 3: Critical 
    \begin{itemize}
        \item Pixels of a darker and/or duller blue which have not been classified as containing active fire, may be considered to be false negatives. This distinction depends mainly on the brightness of the blue colour, and in this approach even a very dark/dull blue is considered a false negative. 
        \item Areas of pixels of a middle and dark shade blue colour are counted as discrete active fires if they are distant or close to another detection, or another ambiguous fire. 
        \item Pixels of a middle shade blue may be considered a false detection if the area is small and distant from other detections. 
    \end{itemize}
\end{itemize}
 
For each of the three approaches, an absolute value for false positives and false negatives was calculated.  To calculate the false positives as a percentage of all detections, their number was divided by all the discrete detections made during the pass, which was 921. To calculate the false negatives as a percentage of all detections, the number of false negatives was divided by the sum of all the discrete detections made during the pass and the false negatives. The results are summarised in table \ref{tab:internal}

\begin{table}[h]
\caption{\label{tab:internal}Integration test results}
\begin{center}
\begin{tabular}{|l|c|c|c|c|}
\hline
\rowcolor[HTML]{C0C0C0} 
\textbf{Approach} & \multicolumn{1}{l|}{\cellcolor[HTML]{C0C0C0}\textbf{False -}} & \multicolumn{1}{l|}{\cellcolor[HTML]{C0C0C0}\textbf{False +}} & \multicolumn{1}{l|}{\cellcolor[HTML]{C0C0C0}\textbf{\% False -}} & \multicolumn{1}{l|}{\cellcolor[HTML]{C0C0C0}\textbf{\% False +}} \\ \hline
Generous          & 0                                                             & 0                                                             & 0                                                                & 0                                                                \\ \hline
Moderate          & 4                                                             & 27                                                            & 0.43                                                             & 2.85                                                             \\ \hline
Critical          & 7                                                             & 96                                                            & 0.76                                                             & 9.44                                                             \\ \hline
\end{tabular}
\end{center}
\end{table}

The results indicate that false negatives are calculated to be a maximum of 0.76\% from the integration tests. This satisfies the safety requirement to identify 95\% of all active wildfires in the area of deployment. The safety requirement for a maximum of 52 false positive detections per month was marginally missed using critical validation approach but was met using the moderate and generous approaches.

\section{Conclusions and Future Work}\label{sec:conc}

In this paper we have described the application of a safety assurance process to a machine learned satellite-based wildfire detection and alert component and shown how a compelling safety case for the component was created as the output of that process. The process applied was the AMLAS approach \cite{hawkins2021guidance} consisting of 6 steps, each of which generated part of the safety argument for the ML component. Each of these fragments of safety argument presented in this paper are connected together to provide the complete safety argument and evidence for the ML safety case. This ML safety case is then integrated as part of the overall safety case for wildfire alert system. This overall safety case also considers the assurance of other elements of the wildfire alert system such as the satellite, the communications and the fire service response.

As far as we are aware, the work presented in this paper represents the first fully developed safety case for an ML component containing explicit argument and evidence as to the safety of the ML. We intend to develop further the deployment aspects of the safety case, once the development of the system moves further into the deployment phase. In addition, we will extend this work to consider operational changes and updates and the impact that these have on the validity of the safety case during operation.

\section*{Acknowledgement}
The work is funded by the  Assuring Autonomy International Programme (www.york.ac.uk/assuring-autonomy).

\bibliographystyle{ieeetr}
\bibliography{actions}

\section{Statements and Declarations}

\subsection{Funding}

The work is funded by the  Assuring Autonomy International Programme (www.york.ac.uk/assuring-autonomy).

\subsection{Competing Interests}

The authors have no relevant financial or non-financial interests to disclose.

\subsection{Author Contributions}

All authors contributed to the work reported in this paper. The first draft of the manuscript was written by Richard Hawkins and all authors commented on previous versions of the manuscript. All authors read and approved the final manuscript.

\subsection{Data Availability}

The datasets generated during and/or analysed during the current study are available from the corresponding author on reasonable request.

\subsection{Ethics Approval}

The research reported in this paper does not involve human or animal subjects.

\subsection{Consent to participate}

The research reported in this paper does not involve human subjects.

\subsection{Consent for publication}

The research reported in this paper does not involve human subjects.

\end{document}